\documentclass{article}

\usepackage{nips11submit_e,times}

\usepackage{times}
\usepackage{epsfig}
\usepackage{subfigure}
\usepackage{graphicx}
\usepackage{amsmath}
\usepackage{amssymb}
\usepackage{verbatim}

\usepackage[breaklinks=true,letterpaper=true,colorlinks,bookmarks=false]{hyperref}
\nipsfinalcopy

\begin{document}

%%%%%%%%% TITLE
\title{Deep Convolutional Ranking for\\Multilabel Image Annotation}

\author{
Yunchao Gong\\
UNC Chapel Hill\\
\footnotesize{\texttt{yunchao@cs.unc.edu}} \\
\And
Yangqing Jia\\
Google Research\\
\footnotesize{\texttt{jiayq@google.com}} \\
\And
Thomas K. Leung\\
Google Research\\
\footnotesize{\texttt{leungt@google.com}} \\
\And
Alexander Toshev\\
Google Research\\
\footnotesize{\texttt{toshev@google.com}} \\
\And
Sergey Ioffe\\
Google Research\\
\footnotesize{\texttt{sioffe@google.com}} \\
}

\maketitle

%%%%%%%%% ABSTRACT

\maketitle
\begin{abstract}
Multilabel image annotation is one of the most important challenges in computer vision with many real-world applications. While existing work usually use conventional visual features for multilabel annotation, features based on Deep Neural Networks have shown potential to significantly boost performance. In this work, we propose to leverage the advantage of such features and analyze key components that lead to better performances. Specifically, we show that a significant performance gain could be obtained by combining convolutional architectures with approximate top-$k$ ranking objectives, as thye naturally fit the multilabel tagging problem. Our experiments on the NUS-WIDE dataset outperforms the conventional visual features by about 10\%, obtaining the best reported performance in the literature.
\end{abstract}

\section{Introduction}

\begin{figure}[]
\centering
   \includegraphics[width=5in,  trim= 35mm 80mm 40mm 70mm]{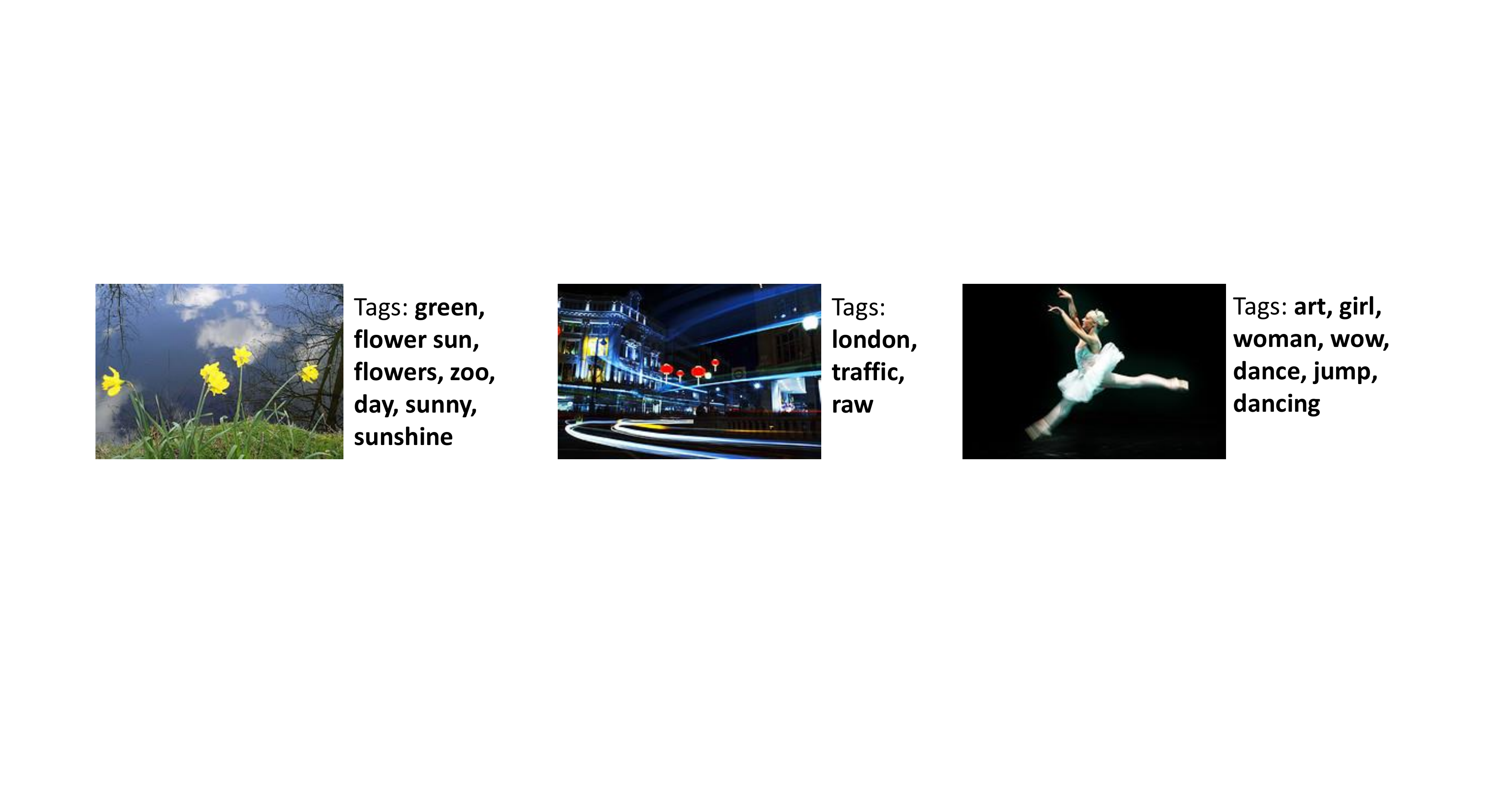}
	\caption{Sample images from the NUS-WIDE dataset, where each image is annotated with several tags.
	\label{sample}}
\end{figure}

Multilabel image annotation \cite{Makadia08,Guillaumin09} is an important and challenging problem in computer vision. Most existing work focus on single-label classification problems \cite{imagenet,lazebnik06}, where each image is assumed to have only one class label.  However, this is not necessarily true for real world applications, as an image may be associated with multiple semantic tags (Figure \ref{sample}). As a practical example, images from Flickr are usually accompanied by several tags to describe the content of the image, such as objects, activities, and scene descriptions. Images on the Internet, in general, are usually associated with sentences or descriptions, instead of a single class label, which may be deemed as a type of multitagging. Therefore, it is a practical and important problem to accurately assign multiple labels to one image.

Single-label image classification  has been extensively studied in the vision community, the most recent advances reported on the large-scale ImageNet database \cite{imagenet}. Most existing work focus on designing visual features for improving recognition accuracy. For example, sparse coding \cite{wangjj10,yang09}, Fisher vectors \cite{Perronnin07}, and VLAD \cite{Jegou2010} have been proposed to reduce the quantization error of ``bag of words''-type features. Spatial pyramid matching \cite{lazebnik06} has been developed to encode spatial information for recognition. Very recently, deep convolutional neural networks (CNN) have demonstrated promising results for single-label image classification \cite{cnn}. Such algorithms have all focused on learning a better feature representation for one-vs-rest classification problems, and it is not yet clear how to best train an architecture for multilabel annotation problems.

In this work, we are interested in leveraging the highly expressive convolutional network for the problem of multilabel image annotation. We employed a similar network structure to \cite{cnn}, which contains several convolutional and dense connected layers as the basic architecture. We studied and compared several other popular multilabel losses, such as the ranking loss \cite{Joachims} that optimizes the area under ROC curve (AUC), and the cross-entropy loss used in Tagprop \cite{Guillaumin09}. Specifically, we propose to use the top-$k$ ranking loss, inspired by \cite{Weston11}, for embedding to train the network. Using the largest publicly available multilabel dataset NUS-WIDE \cite{nus-wide-civr09}, we observe a significant performance boost over conventional features, reporting the best retrieval performance on the benchmark dataset in the literature.

\subsection{Previous Work\label{previous}}

In this section we  first review related works on multilabel image annotation and then briefly discuss works on deep convolutional networks.

Modeling Internet images and their corresponding textural information (e.g., sentences, tags) have been of great interest in the vision community \cite{berg07,fergus09,gong11,guillaumin10,Rasiwasia07,Quattoni07,Weston11}. In this work, we focus on the image annotation problem and summarize several important lines of related research. Early work in this area was mostly devoted to annotation models inspired by machine translation techniques \cite{Barnard01,Duygulu02}. The work by Barnard et al. \cite{Barnard01,Duygulu02}  applied machine translation methods to parse natural images and tried to establish a relationship between image regions and words.

More recently, image annotation has been formulated as a classification problem. Early works focused on generative model based tagging \cite{Barnard01,Carneiro2007,Monay04}, centred upon the idea of learning a parametric model to perform predictions. However, because image annotation is a highly nonlinear problem, a parametric model might not be sufficient to capture the complex distribution of the data. Several recent works on image tagging have mostly focused on nonparametric nearest-neighbor methods, which offer higher expressive power. The work by Makadia et al. \cite{Makadia08}, which proposed a simple nearest-neighbor-based tag transfer approach, achieved significant improvement over previous model-based methods. Recent improvements on the nonparametric approach include TagProp \cite{Guillaumin09}, which learns a discriminative metric for nearest neighbors to improve tagging.

Convolutional neural networks (CNNs) \cite{cnn,lecun90,lee09,jarrett09,NIPS2012_0598} are a special type of neural network that utilizes specific network structures, such as convolutions and spatial pooling, and have exhibited good generalization power in image-related applications. Combined with recent techniques such as Dropout and fast parallel training, CNN models have outperformed existing hancrafted features. Krizhevsky et al. \cite{cnn} reported record-breaking results on ILSVRC 2012 that contains 1000 visual-object categories. However, this study was mostly concerned with single-label image classification, and the images in the dataset only contain one prominent object class. At a finer scale, several methods focus on improving specific network designs. Notably, Zeiler et al. \cite{zeiler13} investigated different pooling methods for training CNNs, and several different regularization methods, such as Dropout \cite{Hinton12}, DropConnect \cite{wan13}, and Maxout \cite{goodfellow13} have been proposed to improve the robustness and representation power of the networks. In adition, Earlier studies \cite{donahue2013decaf} have shown that CNN features are suitable as a general feature for various tasks under the conventional classification schemes, and our work focuses on how to directly train a deep network from raw pixels, using multilabel ranking loss, to address the multilabel annotation problem.

\section{Multilabel Deep Convolutional Ranking Net}

In our approach for multilabel image annotation, we adopted the architecture proposed in \cite{cnn} as our basic framework and mainly focused on training the network with loss functions tailored for multi-label prediction tasks.

\subsection{Network Architecture}

The basic architecture of the network we use is similar to the one used in \cite{cnn}. We use five convolutional layers and three densely connected layers. Before feeding the images to the convolutional layers, each image is resized to 256$\times$256. Next, 220$\times$220 patches are extracted from the whole image, at the center and the four corners to provide an augmentation of the dataset. 
Convolution filter sizes are set to squares of size 11, 9, and 5 respectively for the different convolutional layers; and max pooling layers are used in some of the convolutional layers to introduce invariance. Each densely connected layer has  output sizes of 4096. Dropout layers follow each of the densely connected layers  with a dropout ratio of 0.6. For all the layers, we used rectified linear units (RELU) as our nonlinear activation function.

The optimization of the whole network is achieved by asynchronized stochastic gradient descent with a momentum term with weight $0.9$, with mini-batch size of 32. The global learning rate for the whole network is set to $0.002$ at the beginning, and a staircase weight decay is applied after a few epochs. The same optimization parameters and procedure are applied to all the different methods. For our dataset with 150,000 training images, it usually takes one day to obtain a good model by training on a cluster. Unlike previous work that usually used ImageNet to pre-train the network, we train the whole network directly from the training images from the NUS-WIDE dataset for a fair comparison with conventional vision baselines.

\subsection{Multilabel Ranking Losses}

We mainly focused on loss layer, which specifies how the network training penalizes the deviation between the predicted and true labels, and investigated several different multilabel loss functions for training the network. The first loss function was inspired by Tagprop \cite{Guillaumin09}, for which we minimized the multilabel softmax regression loss. The second loss was a simple modification of a pairwise-ranking loss \cite{Joachims}, which takes multiple labels into account. The third loss function was a multilabel variant of the WARP loss \cite{Weston11}, which uses a sampling trick to optimize top-$k$ annotation accuracy.

For notations, assume that we have a set of images $\boldsymbol x$ and that we denote the convolutional network by $f(\cdot)$ where the convolutional layers and dense connected layers filter the images. The output of $f(\cdot)$ is a scoring function of the data point $\boldsymbol x$, that produces a vector of activations.  We assume there are $n$ image training data and $c$ tags.

\subsubsection{Softmax}

The softmax loss has been used for multilabel annotation in Tagprop \cite{Guillaumin09}, and is also used in single-label image classification \cite{cnn}; therefore, we adopted it in our context. The posterior probability of an image $\boldsymbol x_i$ and class $j$ can be expressed as
\begin{equation}
p_{ij} = \frac{\exp(f_j(\boldsymbol x_i))}{ \sum_{k=1}^c \exp(f_k(\boldsymbol x_i))},
\end{equation}
where $f_j(\boldsymbol x_i)$ means the activation value for image $\boldsymbol x_i$ and class $j$. We then minimized the KL-Divergence between the predictions and the ground-truth probabilities. Assuming that each image has multiple labels, and that we can form a label vector $\boldsymbol y \in R^{1 \times c}$ where $y_j = 1$ means the presence of a label and $y_j = 0$ means absence of a label for an image, we can obtain ground-truth probability by normalizing $\boldsymbol y$ as $\boldsymbol y / \|\boldsymbol y \|_1$. If the ground truth probability for image $i$ and class $j$ is defined as $\bar{p}_{ij}$, the cost function to be minimized is
\begin{equation}
J = -\frac{1}{m} \sum_{i=1}^n \sum_{j=1}^{c} \bar{p}_{ij} \log(p_{ij}) = -\frac{1}{m} \sum_{i=1}^n \sum_{j=1}^{c_+} \frac{1}{c_+}\log(p_{ij}) \nonumber
\end{equation}
where $c_+$ denotes the number of positive labels for each image. For the ease of exposition and without loss of generality, we set $c_+$ to be the same for all images.

\subsubsection{Pairwise Ranking}

The second loss function we considered was the pairwise-ranking loss \cite{Joachims}, which directly models the annotation problem. In particular, we wanted to rank the positive labels to always have higher scores than negative labels, which led to the following minimization problem:
\begin{equation}
J = \sum_{i=1}^n \sum_{j=1}^{c_+} \sum_{k=1}^{c_{-}} \max(0, 1 - f_{j}( \boldsymbol  x_i) + f_{k}(\boldsymbol  x_i)),
\end{equation}
where $c_+$ is the positive labels and  $c_{-}$ is the negative labels. During the back-propagation, we computed the sub-gradient of this loss function. One limitation of this loss is that it optimizes the area under the ROC curve (AUC) but does not directly optimize the top-$k$ annotation accuracy. Because for image annotation problems we were mostly interested in top-$k$ annotations, this pairwise ranking loss did not best fit our purpose.

\subsubsection{Weighted Approximate Ranking (WARP)}

The third loss we considered was the weighted approximate ranking (WARP), which was first described in \cite{Weston11}. It specifically optimizes the top-$k$ accuracy for annotation by using a stochastic sampling approach. Such an approach fits the stochastic optimization framework of the deep architecture very well. It minimizes
\begin{equation}
J =  \sum_{i=1}^n \sum_{j=1}^{c_+} \sum_{k=1}^{c_{-}} L(r_j) \max(0, 1 - f_{j}(\boldsymbol x_i) + f_{k}(\boldsymbol x_i)).
\end{equation}
where $L(\cdot)$ is a weighting function for different ranks, and $r_j$ is the rank for the $j$th class for image $i$. The weighting function $L(\cdot)$ used in our work is defined as:
\begin{equation}
L(r) = \sum_{j=1}^{r} \alpha_j, ~\text{with~} \alpha_1 \geq \alpha_2 \geq \ldots  \geq 0.
\end{equation}
We defined the $\alpha_i$ as equal to $1/j$, which is the same as \cite{Weston11}. The weights defined by $L(\cdot)$ control the top-$k$ of the optimization. In particular, if a positive label is ranked top in the label list, then $L(\cdot)$ will assign a small weight to the loss and will not cost the loss too much. However, if a positive label is not ranked top, $L(\cdot)$ will assign a much larger weight to the loss, which pushes the positive label to the top. The last question was how to estimate the rank $r_j$ . We followed the sampling method in \cite{Weston11}: for a positive label, we continued to sample negative labels until we found a violation; then we recorded the number of trials s we sampled for negative labels. The rank was estimated by the following formulation
\begin{equation}
r_j = \lfloor \frac{c-1}{s} \rfloor,
\end{equation}
for $c$ classes and $s$ sampling trials. We computed the sub-gradient for this layer during optimization.

As a minor noite, the approximate objective we optimize is a looser upper bound compared to the original WARP loss proposed in \cite{Weston11}. To see this, notice that in the original paper, it is assumed that the probability of sampling a violator is $p = \frac{r}{\#Y - 1}$ with a positive example $(x,y)$ with rank $r$, where $\#Y$ is the number of labels. Thus, there are $r$ labels with higher scores than $y$. This is true only if all these $r$ labels are negative. However, in our case, since there may be other positive labels having higher scores than $y$ due to the multi-label nature of the problem, we effectively have $p \leq \frac{r}{\#Y - 1}$.

\section{Visual Feature based Image Annotation Baslines}

We used a set of 9 different visual features and combined them to serve as our baseline features. Although such a set of features might not have been the best possible ones we could obtain, they already serve as a very strong visual representation, and the computation of such features is nontrivial. On top of these features, we ran two simple but powerful classifiers (kNN and SVM) for image annotation. We also experimented with Tagprop \cite{Guillaumin09}, but found it cannot easily scale to a large training set because of the $O(n^2)$ time complexity. After using a small training set to train the Tagprop model, we found the performance to be unsatisfactory and therefore do not compare it here.

\subsection{Visual Features}

\noindent\textbf{GIST} \cite{gist}: We resized each image to 200$\times$200 and used three different scales [8, 8, 4] to filter each
RGB channel, resulting in 960-dimensional (320$\times$3) GIST feature vectors.\smallskip

\noindent\textbf{SIFT}: We used two different sampling methods and three different local descriptors to extract texture features, which gave us a total of 6 different features. We used dense sampling and a Harris corner detector as our patch-sampling methods. For local descriptors, we extracted SIFT \cite{lowe04}, CSIFT \cite{Sande10}, and RGBSIFT \cite{Sande10}, and formed a codebook of size 1000 using kmeans clustering; then built a two-level spatial pyramid \cite{lazebnik06} that resulted in a 5000-dimensional vector for each image. We will refer to these six features as D-SIFT, D-CSIFT, D-RGBSIFT, H-SIFT, H-CSIFT, and H-RGBSIFT.\smallskip

\noindent\textbf{HOG}: To represent texture information at a larger scale, we used 2$\times$2 overlapping HOG as described in \cite{sun}. We quantized the HOG features to a codebook of size 1000 and used the same spatial pyramid scheme as above, which resulted in 5000-dimensional feature vectors.\smallskip

\noindent\textbf{Color}: We used a joint RGB color histogram of 8 bins per dimension, for a 512-dimensional feature.\smallskip

 The same set of features were used in \cite{gong13}, and achieved state-of-the-art performance for image retrieval and annotation.
The combination of this set of features has a total dimensionality of 36,472, which makes learning  very expensive. We followed \cite{gong13} to perform simple dimensionality reductions to reduce computation. In particular, we performed a kernel PCA (KPCA) separately on each feature to reduce the dimensionality to 500. Then we concatenated all of the feature vectors to form a 4500-dimensional global image feature vector and performed different learning algorithms on it.

\subsection{Visual feature + $k$NN}

The simplest baseline that remains very powerful involves directly applying a weighted kNN on the visual feature vectors. $k$NN is a very strong baseline for image annotation, as suggested by Makadia et al. \cite{Makadia08}, mainly because multilabel image annotation is a highly nonlinear problem and handling the heavily tailed label distribution is usually very difficult. By contrast, kNN is a highly nonlinear and adaptive algorithm that better handles rare tags. For each test image, we found its $k$ nearest neighbors in the training set and computed the posterior probability $p(c|i)$ as
\begin{equation}
p(c|i) = \sum_{j=1}^k  \frac{1}{k} \exp({-\frac{||x_i - x_j||_2^2}{\sigma}}) y_{jk},
\end{equation}
where $y_{ik}$ indexes the labels of training data, $y_{ik}=1$ when there is one label for this image, and $y_{ik}=0$ when there is no label for this image. $\sigma$ is the bandwidth that needs to be tuned. After obtaining the prediction probabilities for each image, we sorted the scores and annotated each testing image with the top-$k$ tags.

\subsection{Visual feature + SVM}

Another way to perform image annotation is to treat each tag separately and to train $c$ different one-vs-all classifiers. We trained a linear SVM \cite{liblinear} for each tag and used the output of the $c$ different SVMs to rank the tags. Because we had already performed nonlinear mapping to the data during the KPCA stage, we found a linear SVM to be sufficient. Thus we assigned top-$k$ tags to one image, based on the ranking of the output scores of the SVMs.

\section{Experiments\label{sec:exp}}

\subsection{Dataset} \label{sec:datasets}

We performed experiments on the largest publicly available multilabel dataset, NUS-WIDE \cite{nus-wide-civr09}. This dataset contains 269,648 images downloaded from Flickr that have been manually annotated, with several tags (2-5 on average) per image. After ignoring the small subset of the images that are not annotated by any tag, we had a total of 209,347 images for training and testing. We used a subset of 150,000 images for training and used the rest of the images for testing. The tag dictionary for the images contains 81 different tags. Some sample images and annotations  are shown in Figure \ref{sample}.

\subsection{Evaluation Protocols}

We followed previous research \cite{Makadia08} in our use of the following protocols to evaluate different methods. For each image, we assigned $k$ (e.g., $k = 3, 5$) highest-ranked tags to the image and compared the assigned
tags to the ground-truth tags. We computed the recall and precision for each tag separately, and then report the mean-per-class recall and mean-per-class precision:
\begin{equation}
\text{per-class~recall} = \frac{1}{c} \sum_{i=1}^c \frac{N_i^c}{N_i^g},~~~~~\text{per-class~precision}  = \frac{1}{c} \sum_{i=1}^c \frac{N_i^c}{N_i^p}\\
\end{equation}
where $c$ is the number of tags, $N_i^c$  is the number of correctly annotated image for tag $i$, $N_i^g$ is the number of ground-truth tags for tag $i$, and $N_i^p$ is the number of predictions for tag $i$. The above evaluations are biased toward infrequent tags, because making them correct would have a very significant impact on the final accuracy. Therefore we also report the overall recall and overall precision:
\begin{equation}
\text{overall~recall} =  \frac{ \sum_{i=1}^c N_i^c}{ \sum_{i=1}^c N_i^g},~~~~~\text{overall~precision}  =   \frac{ \sum_{i=1}^c N_i^c}{ \sum_{i=1}^c N_i^p}.\\
\end{equation}
For the above two metrics, the frequent classes will be dominant and have a larger impact on final performance. Finally, we also report the percentage of recalled tag words out of all tag words as N+. We believe that evaluating all of these metrics makes the evaluation unbiased.

\subsection{Baseline Parameters}

In our preliminary evaluation, we optimized the parameters for the visual-feature-based baseline systems. For visual-feature dimensionality reduction, we followed the suggestions in Gong et al. \cite{gong13}  to reduce the dimensionality of each feature to 500 and then concatenated the PCA-reduced vectors into a 4500-dimensional global image descriptor, which worked as well as the original feature. For $k$NN, we set the bandwidth $\sigma$ to 1 and $k$ to 50, having found that these settings work best. For SVM, we set the regularization parameter to $C=2$, which works best for this dataset.

\begin{table*}[]
\begin{center}\small
\setlength{\tabcolsep}{0.15cm}
\begin{tabular}{c|cccccc}
\hline
method / metric    &  per-class recall &      per-class precision 		& overall recall		&	overall precision    &		$N+$        \\ \hline
  Upper bound    &       	 97.00		&	 44.87		&	 82.76	&	 66.49 	&	100.00      \\ \hline
  Visual Feature + kNN		&	19.33 	&	\textbf{32.59} 	&	53.44 	&	42.93 	&	91.36	\\
    Visual Feature + SVM	&		18.79 	&	21.51 	&	35.87 	&	28.82 	&	82.72	\\ \hline
  CNN + Softmax		&	31.22 	&	31.68	&	 59.52	&	 47.82	&	 98.76	\\
  CNN + Ranking		&	26.83	&	 31.93 	&	58.00 	&	46.59 	&	95.06	\\
  CNN + WARP			&	\textbf{35.60} 	&	{31.65}	&	 \textbf{60.49}	&	 \textbf{48.59}	&	 \textbf{96.29}  \\
  \hline
\end{tabular}
\caption{Image annotation results on NUS-WIDE with $k=3$ annotated tags per image. See text in section 5.4 for the definition of ``Upper bound''.
\label{knn3}}
\end{center}
\end{table*}

\begin{table*}[]
\begin{center}\small
\setlength{\tabcolsep}{0.15cm}
\begin{tabular}{c|cccccc}
\hline
method / metric    &  per-class recall &      per-class precision 		& overall recall		&	overall precision    &		$N+$        \\ \hline
  Upper bound    &       	99.57	&	 28.83 	&	96.40 	&	46.22	&	 100.00        \\ \hline
  Visual Feature + kNN	&      32.14 	&	\textbf{22.56} 	&	66.98 	&	32.29	&	 95.06	\\
    Visual Feature + SVM	&	34.19 	&	18.79	&	 47.15 	&	22.73 	&	96.30	\\ \hline
  CNN + Softmax		&	48.24	&	 21.98	&	 74.04	&	 35.69	&	 98.76	\\
  CNN + Ranking		&	42.48 	&	22.74 	&	72.78 	&	35.08 	&	97.53	\\
  CNN + WARP			&	\textbf{52.03} 	&	{22.31}	&	 \textbf{75.00}	&	 \textbf{36.16}	&	 \textbf{100.00}  \\
  \hline
\end{tabular}
\caption{Image annotation results on NUS-WIDE with $k=5$ annotated tags per image. See text in section 5.4 for the definition of ``Upper bound''. \label{knn5}}
\end{center}
\end{table*}

\subsection{Results}

We first report results with respect to the metrics introduced above. In particular, we vary the number $k$ of predicted keywords for each image and mainly consider $k = 3$ and $k = 5$. Before doing so, however, we must define an upper bound for our evaluation. In the dataset, each image had different numbers of ground-truth tags, which made it hard for us to precisely compute an upper bound for performance with different $k$. For each image, when the number of ground-truth tags was larger than $k$, we randomly chose $k$ ground-truth tags and assigned them to that image; when the number of ground-truth tags was smaller than $k$, we assigned all ground-truth tags to that image and randomly chose other tags for that image. We believe this baseline represents the best possible performance when the ground truth is known. The results for assigning 3 keywords per image are reported in Table \ref{knn3}. The results indicate that the deep network achieves a substantial improvement over existing visual-feature-based annotation methods. The CNN+Softmax method outperforms the VisualFeature+SVM baseline by about 10\%.
Comparing the same CNN network with different loss functions, results show that softmax already gives a very powerful baseline. Although using the pairwise ranking loss does not improve softmax, by using the weighted approximated-ranking loss (WARP) we were able to achieve a substantial improvement over softmax. This is probably because pairwise-ranking is not directly optimizing the top-$k$ accuracy,  and because WARP pushes classes that are not ranked top heavier, which boosts the performance of rare tags. From these results, we can see that all loss functions achieved comparable overall-recall and overall-precision, but that WARP loss achieved significantly better per-class recall and per-class precision. Results for $k = 5$, which are given in Table \ref{knn5}, show similar trends to $k = 3$.

\begin{figure*}[]
\centering
   \includegraphics[width=4.5in, trim= 0mm 110mm 0mm 120mm]{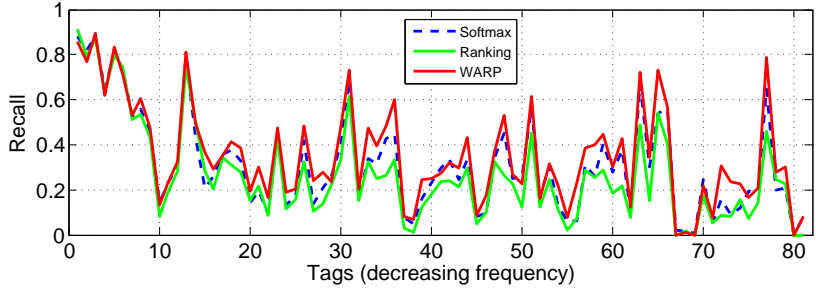}
	\caption{Analysis of per-class recall of the 81 tags in NUS-WIDE dataset with $k=3$.\label{per-class1}}
\end{figure*}

\begin{figure*}[]
\centering
   \includegraphics[width=4.5in, trim= 0mm 110mm 0mm 110mm]{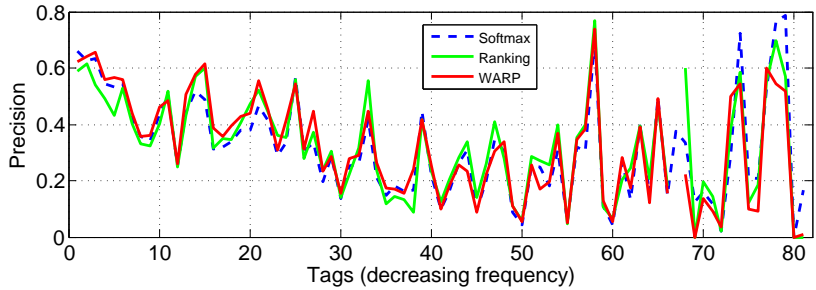}
	\caption{Analysis of per-class precision of the 81 tags in NUS-WIDE dataset with $k=3$.\label{per-class2}}
\end{figure*}

\begin{figure*}[]
\centering
   \includegraphics[width=4.5in, trim= 5mm 10mm 5mm 10mm]{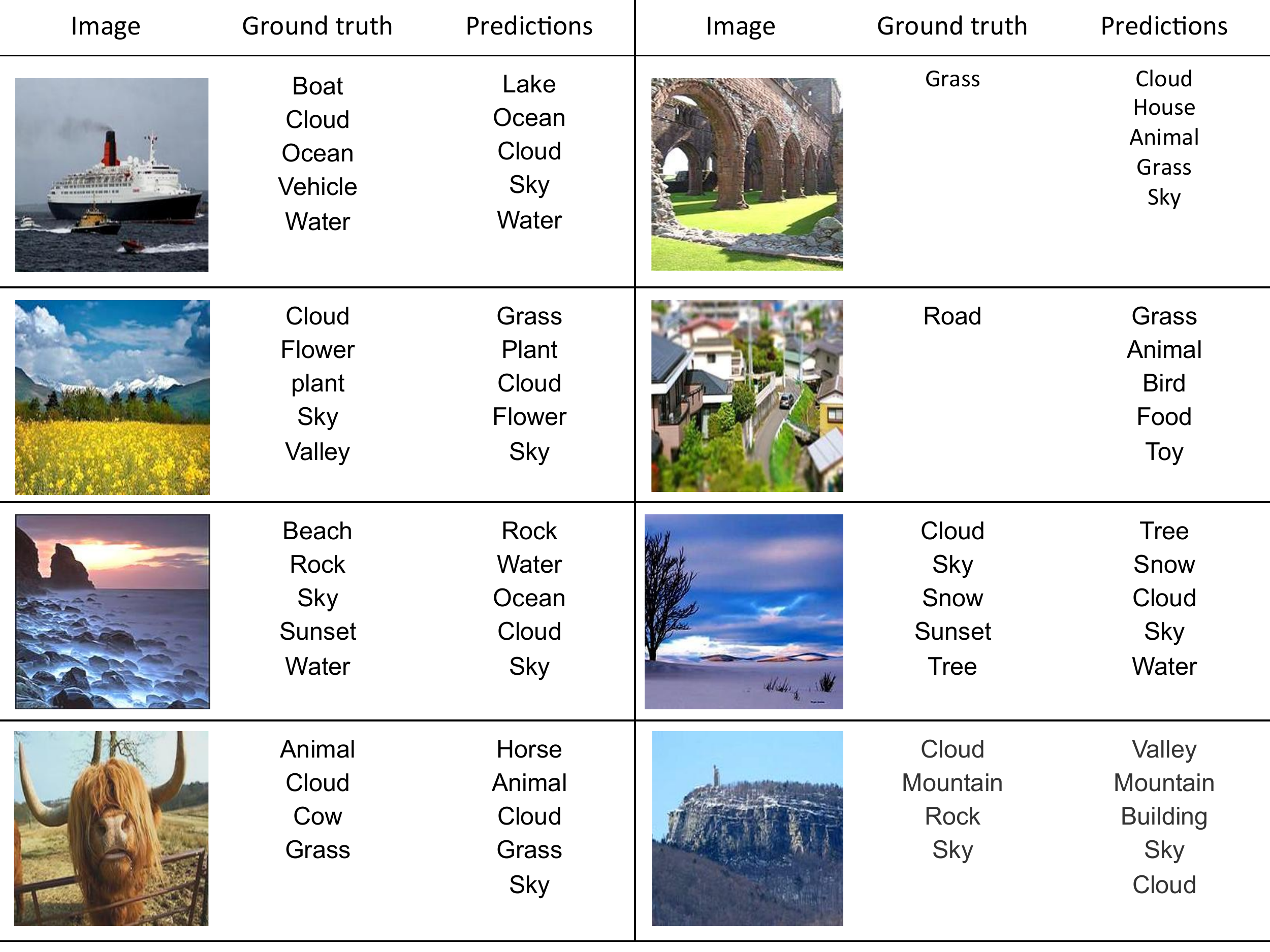}
	\caption{Qualitative image annotation results obtained with WARP.\label{anno}}
\end{figure*}

We also provide a more detailed analysis of per-class recall and per-class precision. The recall for each tags appears in Figure \ref{per-class1}, and the precision for each tag in Figure \ref{per-class2}. The results for different tags are sorted by the frequency of each tag, in descending order. From these results, we see that the accuracy for frequent tags greater than for infrequent tags. Different losses performed comparably to each other for frequent classes, and WARP worked better than other loss functions for infrequent classes. Finally, we show some image annotation examples in Figure \ref{anno}. Even though some of the predicted tags for these do not match the ground truth, they are still very meaningful.

\section{Discussion and Future Work}

In this work, we proposed to use ranking to train deep convolutional neural networks for multilabel image annotation problems. We investigated several different ranking-based loss functions for training the CNN, and found that the weighted approximated-ranking loss works particularly well for multilabel annotation problems. We performed experiments on the largest publicly available multilabel image dataset NUS-WIDE, and demonstrated the effectiveness of using top-$k$ ranking to train the network. In the future, we would like to use very large amount of noisy-labeled multilabel images from the Internet (e.g., from Flickr or image searches) to train the network.

%{
%\bibliographystyle{plain}
%\bibliography{egbib}
%}
% for arxiv. Uncomment the above lines and comment the below line if you recompile.

\end{document}